\documentclass{article}

\usepackage{arxiv}

\usepackage[utf8]{inputenc} 
\usepackage[T1]{fontenc}    
\usepackage{hyperref}       
\usepackage{url}            
\usepackage{booktabs}       
\usepackage{amsfonts}       
\usepackage{nicefrac}       
\usepackage{microtype}      
\usepackage{lipsum}
\usepackage{graphicx}

\usepackage{amssymb,amsmath,cases,multirow,epsfig,amsfonts} 
\usepackage[caption=false,font=footnotesize]{subfig}
\usepackage{adjustbox}
\usepackage{graphicx}
\usepackage{dblfloatfix}
\hypersetup{colorlinks = true, linkcolor = blue, pdfauthor=author}
\usepackage[dvipsnames]{xcolor}
\usepackage{mathtools}
\usepackage{booktabs}

\title{Subspace-wise Hybrid RL for Articulated Object Manipulation}

\author{
 Yujin Kim\textsuperscript{1} \footnotemark[1]\\
   \And
 Sol Choi\textsuperscript{2} \\
  \And
 Bum-Jae You\textsuperscript{2} \\
  \And
  Keunwoo Jang\textsuperscript{2} \\
  \And
  Yisoo Lee\textsuperscript{2} \footnotemark[2] \\
}

\begin{document}

\maketitle
\begin{center}
    \textsuperscript{1}Department of Computer Science, Cornell University, USA \\
    \textsuperscript{2}Center for Intelligence \& Interactive Robotics, Korea Institute of Science and Technology (KIST), South Korea \\[1em]
\end{center}

\vspace{1em} %
\renewcommand\thefootnote{} 
\footnotetext{* Now at Cornell University, previously affiliated with KIST.}
\footnotetext{† Corresponding author.}
\footnotetext{This work was supported by the National Research Foundation of Korea (NRF) grant funded by the Korea government(MSIT) (RS-2024-00339632, 2022M3C1A3098746)}
\begin{abstract}
Articulated object manipulation is a challenging task, requiring constrained motion and adaptive control to handle the unknown dynamics of the manipulated objects. While reinforcement learning (RL) has been widely employed to tackle various scenarios and types of articulated objects, the complexity of these tasks, stemming from multiple intertwined objectives makes learning a control policy in the full task space highly difficult. To address this issue, we propose a Subspace-wise hybrid RL (SwRL) framework that learns policies for each divided task space, or subspace, based on independent objectives. This approach enables adaptive force modulation to accommodate the unknown dynamics of objects. Additionally, it effectively leverages the previously underlooked redundant subspace, thereby maximizing the robot's dexterity. Our method enhances both learning efficiency and task execution performance, as validated through simulations and real-world experiments. Supplementary video is available at \href{https://youtu.be/PkNxv0P8Atk}{https://youtu.be/PkNxv0P8Atk}.
\end{abstract}

\section{INTRODUCTION}
Manipulating articulated objects is crucial for enabling robots to operate effectively in human-centered, unstructured environments. Articulated objects, such as doors, valves, and drawers, vary widely in shape and location in real-world settings. Manipulating these objects necessitates constrained motion and precise force modulation, making the task challenging.

Recent studies have attempted to address this problem using learning-based methods \cite{mu2021maniskill,xu2022universal,eisner2022flowbot3d, li2024unidoormanip,beltran2020learning}. These methods are promising due to their ability to generalize across various objects and environments. However, the high degree of freedom and complexity involved in manipulation tasks often result in poor sample efficiency and suboptimal performance when generating full joint or task space control commands. Therefore, decomposing the control space into independent objectives can improve performance, as learning a single objective function is more effective and manageable than learning a multi-objective function.

In the task space, manipulating articulated objects can be decomposed into two subspaces: the geometric constraint subspace and the kinematic constraint subspace \cite{bruyninckx1996specification, berenson2011task}. The subspace associated with maintaining the grasp reflects the geometric shape of the objects and acts as a constraint condition for task execution. Whether the joint of the articulated object is prismatic or revolute, the kinematic constraint subspace exerts force to move the object. 
While these two subspaces can represent the control mechanism of object manipulation, introducing another subspace corresponding to redundancy can further enhance the dexterity of multi-joint robots. 
Some elements related to geometric constraints may not impose actual constraints but instead offer redundancy, allowing the robot to effectively utilize its extra degrees of freedom to optimize the manipulation task, such as avoiding obstacles and minimizing joint stress. 
Although there is no straightforward answer for the optimal motion of these components, effectively utilizing this redundancy can significantly contribute to task success. Despite the effect of the redundant subspace in manipulating articulated objects, no prior works have focused on this aspect. 

We propose a Subspace-wise hybrid Reinforcement Learning(SwRL) framework for articulated object manipulation. We note that the motion in the redundant subspace can enhance task performance in constrained motion scenarios. The RL is employed to generate trajectories for hybrid force/motion control, which is advantageous for contact-rich manipulation. The constrained motion for articulated object manipulation is predefined, while RL is used to determine force modulation and motion in the redundant subspace in real-time during interaction with the object. 

Although data-driven control methods, such as behavior cloning \cite{mu2021maniskill,xu2022universal} and imitation learning\cite{ eisner2022flowbot3d, li2024unidoormanip}, have been extensively studied recently, they cannot effectively utilize redundant subspaces due to the lack of corresponding datasets. In contrast, RL can learn natural motions that maximize the robot's dexterity and perform tasks by utilizing subspaces, which are challenging to engineer manually. Furthermore, we generate adaptive force control profiles that respond to the dynamic properties of the object—such as joint friction, damping, and spring constants—through RL, interacting with the environment. This enables the generation of forces that respond to various physical parameters, similar to those in real-world articulated objects around us.

The SwRL framework imposes independent action spaces on each subspace based on their roles during object manipulation. This approach enhances the interpretability of the objectives, leading to better performance in achieving optimal manipulation motion.

We applied and validated our method on challenging tasks, such as turning valves in industrial settings and manipulating everyday articulated objects like drawers and doors. We validated the effectiveness of our method by applying the trained policies in simulations to a real-world valve rotation task using a Franka Research 3 robot.

\section{RELATED WORKS} \label{sec:relatedworks}
Articulated object manipulation involves adhering to the constrained motions determined by the object's kinematic and geometric characteristics (e.g., sliding prismatic joints, hinging revolute joints). A popular approach to reflect these constraints is path planning-based methods. Given the geometric and kinematic constraints of the object, sampling-based algorithms are used to find the offline path, which is then followed by position controllers \cite{berenson2011task, jaillet2012path, jang2023motion}. The study \cite{berenson2011task} suggests a manipulation planning algorithm under constraints on end-effector pose, while \cite{jaillet2012path} focuses on kinematic constraints, both applying RRT\cite{kuffner2000rrt}. The study \cite{jang2023motion} utilized RRT to find the path of the end-effector in a 2D plane for a door traversal task, while solving inverse kinematics under a fixed pose in 3D space to satisfy constrained motion. While this method works well with precise paths, adaptive controllers \cite{cheah2006adaptive, mittal2022articulated, jain2010pulling,ma2023sim2real} can also be employed to address various uncertainties and the robot’s kinematic and dynamic limitations.

In contact-rich tasks where the robot directly manipulates the object's joints, the uncertainties in control increase due to unobservable dynamics. In such cases, reactive control methods like force/torque control and admittance control are effective. To comply with constrained motion, hybrid force/motion and force/velocity controllers \cite{raibert1981hybrid, mason1981compliance, karayiannidis2016adaptive}  are often utilized. The study \cite{karayiannidis2016adaptive} suggests a hybrid force/velocity controller that estimates the kinematic constraints of an articulated object when its geometric constraints are given. Path planning-based methods, regardless of the controller type, require pre-planned complex object manipulation motions, necessitating substantial engineering effort. This often leads to fixing the end-effector's posture relative to the object during manipulation \cite{berenson2011task,jang2023motion,karayiannidis2016adaptive}, eliminating the potential redundancy space. 

Recent advancements in artificial intelligence have increasingly utilized data-driven learning methods to alleviate the engineering burden of the path-planning process \cite{mu2021maniskill,xu2022universal,eisner2022flowbot3d, li2024unidoormanip,beltran2020learning}. Many studies aim to learn generalized manipulation policies for various objects using approaches like reinforcement learning, imitation learning, and behavior cloning. Imitation learning and behavior cloning, which learn from demonstrations, require high-quality, large-scale demonstration data. These demonstrations often involve heuristic paths, resulting in solutions that limit the multi-joint robot's redundancy \cite{mu2021maniskill,xu2022universal,eisner2022flowbot3d}. Additionally, since these methods learn from demonstrated motions, they may fail to capture the dynamic parameters of the object's joints, such as friction, springs, and dampers, which are common in real-world objects.

While RL shows promising results in adapting to dynamic parameters through environmental interaction, it struggles with the complexity of end-to-end manipulation tasks due to sample efficiency and convergence challenges arising from the large exploration space required \cite{sutton2018reinforcement}. To address these issues, we propose a subspace-wise hybrid RL, force/motion control approach that combines traditional planning-based methods with learning-based techniques. By categorizing the object-oriented task space into three subspaces for training, our method mitigates the challenges associated with large exploration spaces in RL.

\section{PRELIMINARIES}
In this study, articulated objects are manipulated based on the following characteristics:
1) Objects are composed of two or more rigid links connected by one or more revolute or prismatic joints,
2) objects exhibit no displacement or translation other than the sliding or hinging motion of their joints, and
3) objects include geometries that can be defined as handles.

\subsection{Object-Oriented Frame}
\begin{figure}[t!]
    \centering
        \subfloat[\centering {Prismatic Joint}]{
        \includegraphics[width=13mm]{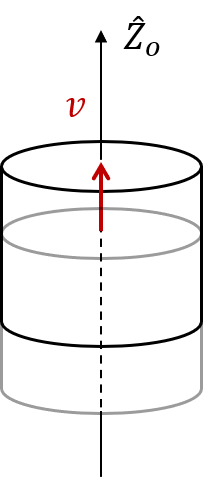}
        \label{fig:prismatic}}
        \hspace{30pt}
        \subfloat[\centering {Revolute Joint}]{
        \includegraphics[width=13mm]{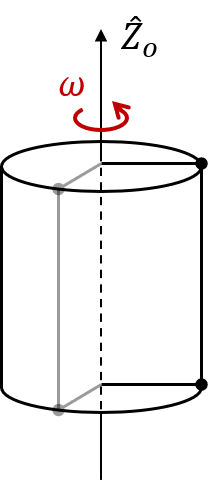}
        \label{fig:revolute}
        }
    \caption{\label{fig:joint}{Illustration of motion directions for prismatic and revolute joints. (a) Prismatic Joint: The motion is linear along the \(\hat{Z}_o\) axis, with velocity \(v\). (b) Revolute Joint: The motion is rotational around the \(\hat{Z}_o\) axis, with angular velocity \(\omega\).}}
\end{figure}

\begin{figure}[t!]
    \centering
        \subfloat[\centering {Door}]{
        \includegraphics[width=.15\textwidth]{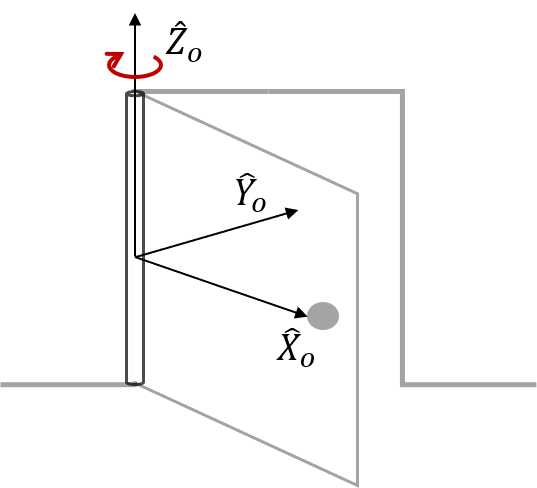}
        \label{fig:door}}
        \hspace{10pt}
        \subfloat[\centering {Valve}]{
        \includegraphics[width=.15\textwidth]{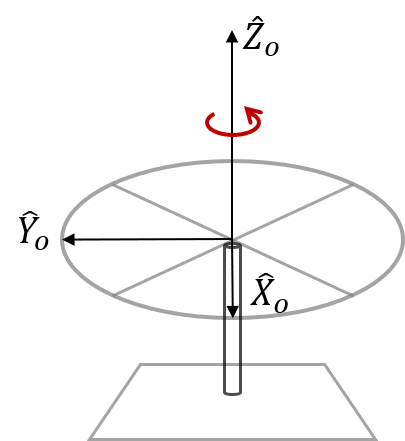}
        \label{fig:valve}
        }
        \hspace{10pt}
        \subfloat[\centering {Drawer}]{
        \includegraphics[width=.15\textwidth]{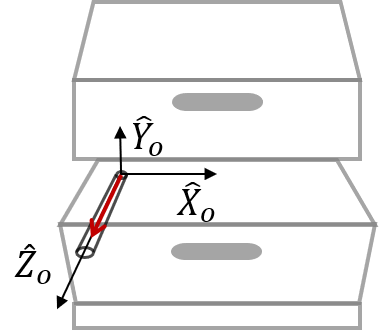}
        \label{fig:drawer}
        }
    \caption{\label{fig:object_frame}{Illustration of object-oriented frame of various objects. }}
\end{figure}

The object-oriented frame $\{O\}$ is a virtual frame located at the joint of the articulated object.
Adopting $\{O\}$ is effective when describing the control component in the task space.
In the frame, the $z$-axis represents the axis of motion as can be seen in Figure~\ref{fig:prismatic},~\ref{fig:revolute}.
The prismatic joint slides along the $z$-axis, and the revolute joint rotates around the $z$-axis.
The $x$-axis is defined as the direction from the joint to the handle position as can be seen in Figure~\ref{fig:object_frame} that depicts the object-oriented frame for various objects.

\begin{figure*}
    \centering
    \includegraphics[width=1\linewidth]{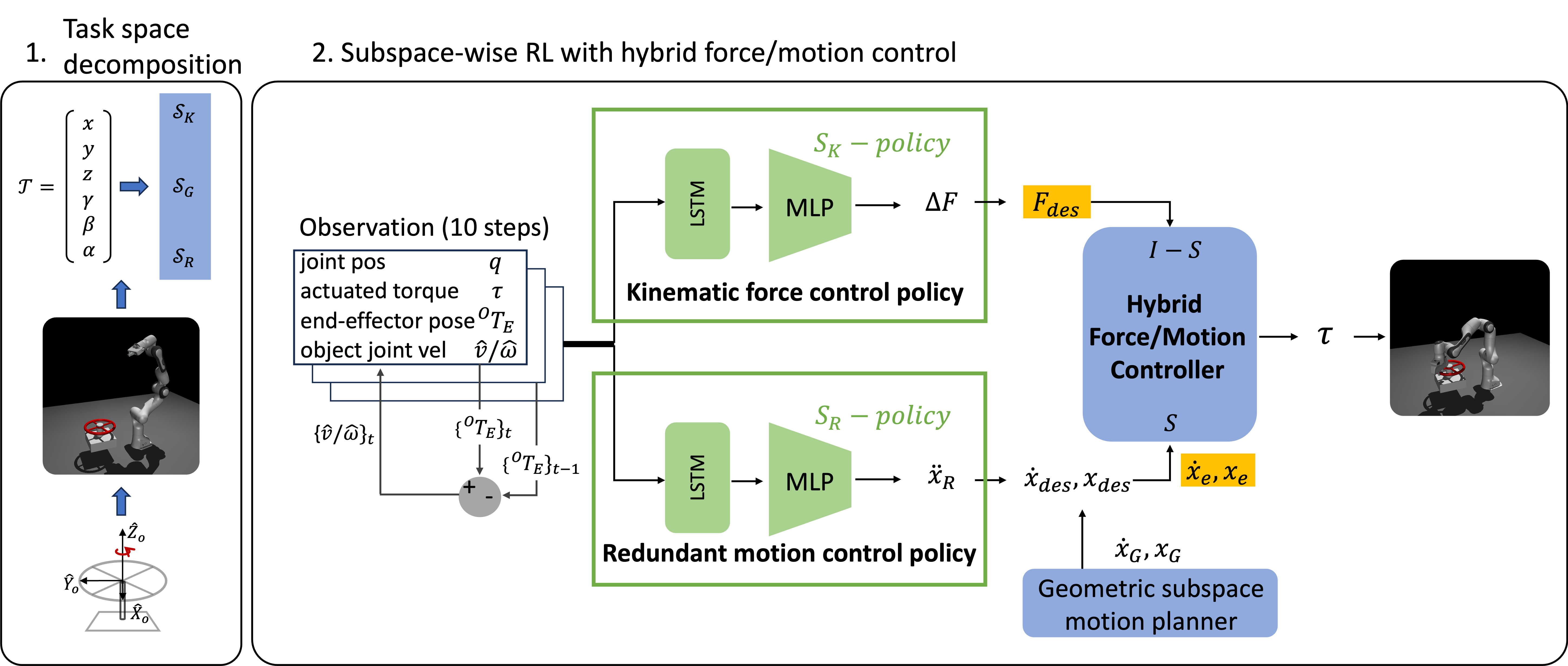}
    \caption{Overview of the Subspace-wise hybrid RL (SwRL) framework for articulated object manipulation. Task space decomposition is performed in the object-oriented frame $\{O\}$, dividing the task space into three subspaces: the kinematic subspace $\mathcal{S}_K$, geometric subspace $\mathcal{S}_G$, and redundant subspace $\mathcal{S}_R$. The \textbf{kinematic force control policy} generates the magnitude of force vector within the kinematic subspace $\mathcal{S}_K$, while the \textbf{redundant motion control policy} produces motion within the redundant subspace $\mathcal{S}_R$. Both policies operate in parallel, with their outputs integrated into the hybrid force/motion controller.}
    \label{fig:framework}
\end{figure*}
\subsection{Hybrid Force/Motion Control}
Hybrid force/motion control was first presented by Raibert et al. in \cite{raibert1981hybrid}. This strategy separates the control of the robot into two orthogonal subspaces: one for position control and the other for force control, allowing the robot to exert precise forces while maintaining accurate positions in specific directions. The equation for hybrid force/motion control is:
\begin{equation}
\begin{aligned}
\tau = J^T\Bigg\{ 
\underbrace{\Lambda S \left(K_p X_e + K_d V_e  \right)}_{\text{motion control}} + \underbrace{(I-S) F_d }_{\text{force control}}
\Bigg\},
\end{aligned}
\label{eq:hybrid_ctrl_equation}
\end{equation}
where \(\tau\) is the vector of joint torques; \(J\) is the Jacobian matrix; \(\Lambda\) is the task space inertia matrix where \(\Lambda=J^{+T}M J^{+}\), $M$ is the joint space inertia matrix, superscript $+$ denotes pseudo inverse; \(S\) is the selection matrix for motion control; \(K_p\) and \(K_d\) are the proportional and derivative gain matrices, respectively; \(X_e\) and \(V_e\) are the position and velocity error vectors in the task space, respectively; and \(F_d\) is the desired force vector.
Force control exerts the desired force in the task space, while motion control follows the operational space control approach as described by \cite{khatib1987unified}. 
Note that all the task space in this study is defined by the object-oriented frame $\{O\}$.

\section{PROPOSED METHOD} \label{sec:method}
Figure~\ref{fig:framework} provides an overview of the Subspace-wise hybrid RL (SwRL) framework.  
In this section, we present our methodology, which integrates a classical controller with reinforcement learning-based control reference generation, leveraging the defined subspaces for object manipulation.  
SwRL is composed of two main components: task space decomposition and subspace-wise RL with hybrid motion and force control.

\subsection{Task Space Decomposition - Three Subspaces} \label{sec:subspace}

In the context of adopting a hybrid force/motion controller, it is essential to select which task space components will be controlled by force and which by motion. When controlling the end-effector in the frame \{O\}, task space can be decomposed into three subspaces based on their roles in manipulating the object. This decomposition forms the basis for selecting the components to be controlled by force or motion.

The task space, \(\mathcal{T} \in \mathbb{R}^6\), represents the configuration space of the end-effector in Cartesian coordinates, encompassing position \([x,y,z]^T\) and orientation \([\gamma, \beta, \alpha]^T\) components in three-dimensional space. We define the task space by three subspaces:
\begin{itemize}
    \item \(\mathcal{S}_K\): The subspace associated with kinematic constraints, which is controlled using force commands. This subspace ensures that the end-effector adheres to the kinematic structure of the object's joint.
    \item \(\mathcal{S}_G\): The subspace associated with geometric constraints, which is controlled using motion commands. This subspace handles constraints related to the geometric configuration of the object.
    \item \(\mathcal{S}_R\): The subspace associated with redundant motion, which can be controlled using either motion or force commands. This subspace addresses the degrees of freedom that are not directly constrained by the task and can be utilized to optimize other criteria, such as avoiding collision or singular pose.
\end{itemize}
The detailed descriptions of each subspace are described in the following.
\subsubsection{ Kinematic Space ($\mathcal{S}_K$)}
$\mathcal{S}_K$ includes components that actuate an articulated object. Adopting force control to these components is advantageous for safe manipulation when the object's dynamics are unknown. This subspace is dependent on the kinematic structure of the object.

Figure~\ref{fig:prismatic} and ~\ref{fig:revolute} illustrate the kinematic structures of prismatic and revolute joints, respectively. Although both objects share the same object-oriented frame and geometric configuration, their kinematic motions differ. For objects with a prismatic joint, motion is constrained along the z-axis, so the kinematic subspace is represented as $\{ z \} \in \mathcal{S}_K$. 

In contrast, for objects with a revolute joint, the kinematic motion occurs as a rotation around the z-axis. While the rotation is centered along the z-axis, the rotational motion can be expressed as a combination of components in the x and y directions at any given moment. Therefore, the kinematic subspace for a revolute joint is defined as $\{ x, y \} \in \mathcal{S}_K$. 

For the valve structure in the handwheel valve in Figure \ref{fig:geom_const}, the kinematic subspace is similarly defined as $\{ x, y\} \in \mathcal{S}_K$, reflecting the rotational motion constraints of the valve handle.

\subsubsection{Geometric Space ($\mathcal{S}_G$)} 
\begin{figure}[htbp]
    \centering
    \begin{minipage}{0.45\textwidth}
        \centering
        \includegraphics[width=0.5\textwidth]{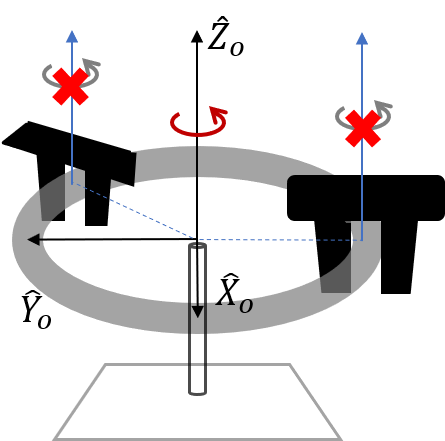}
        \caption{Illustration of geometric constraints for a handwheel valve. The grey circle depicts the handle or the circular trace on the object, while the black object represents the end-effector. The end-effector cannot rotate in the yaw direction from its pose when grasping the object.}
        \label{fig:geom_const}
    \end{minipage}
    \hfill
    \begin{minipage}{0.45\textwidth}
        \centering
        \includegraphics[width=\textwidth]{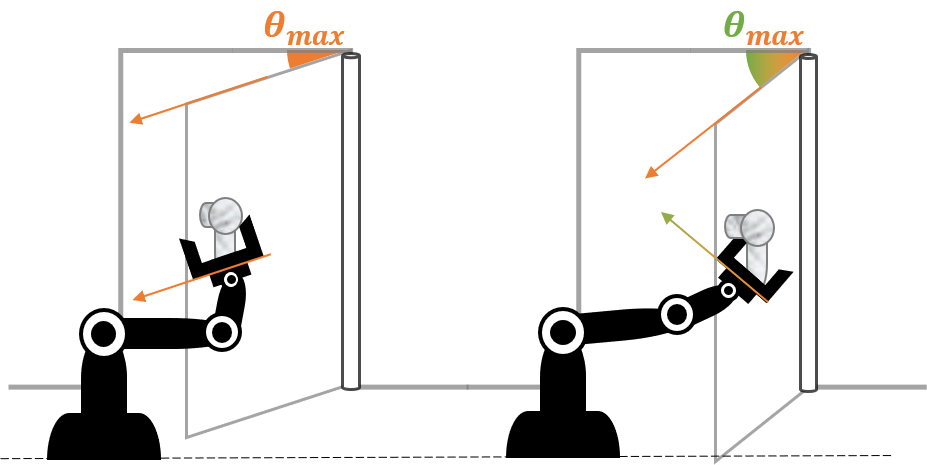}
        \caption{Conceptual illustration of door opening motion with and without redundant space motion. The maximum door opening angle, $\theta_{max}$, is shown for the case without redundant space motion (left) and with redundant motion (right). The right case demonstrates a larger $\theta_{max}$.}
        \label{fig:door_simple}
    \end{minipage}
\end{figure}

$\mathcal{S}_G$ is responsible for ensuring that the robot’s posture satisfies the geometric constraints necessary for manipulating the object. In the example illustrated in Figure~\ref{fig:geom_const}, the yaw angle, $\alpha$, of the gripper is not actively controlled but is passively determined by the relative position of the gripper to the valve on the $x$-$y$ plane. Similarly, the height position, $z$, is deterministic. These constraints ensure that the grasping condition is maintained throughout the motion.

While this setting might appear trivial, it highlights an essential characteristic of the geometric subspace: it encompasses components of the task space that are either predetermined or naturally constrained by the relative pose between the gripper and the object during motion. These components do not need to be learned but can be planned based on the geometry of the object. For the handwheel valve in Figure~\ref{fig:geom_const}, the geometric subspace can be represented as $\{z, \alpha\} \in \mathcal{S}_G$.

\subsubsection{Redundant Space ($\mathcal{S}_R$)} 
While $\mathcal{S}_K$ and $\mathcal{S}_G$ are directly constrained by the object’s configuration, the remaining components of the task space that do not belong to these subspaces can be categorized as redundant elements. These redundant elements are grouped into the redundant subspace, $\mathcal{S}_R$, which provides additional degrees of freedom that can be utilized to optimize the manipulation task. 
The redundant subspace can be expressed as: 
\[
\mathcal{S}_R = \{c \in \mathcal{T} \mid c \notin \mathcal{S}_K, \mathcal{S}_G\}.
\]

Overall, for the handwheel valve shown in Figure~\ref{fig:geom_const}, the task space can be expressed as
\[
\mathcal{T} = \mathcal{S}_K \cup \mathcal{S}_G \cup \mathcal{S}_R, \quad \text{where}
\]
\[
\mathcal{S}_K = \{x, y\}, \quad \mathcal{S}_G = \{z, \alpha\}, \quad \mathcal{S}_R = \{\gamma, \beta\}.
\]

\subsection{Subspace-wise RL with Hybrid Force and Motion Control}
We propose a Subspace-wise hybrid RL(SwRL) framework that learns task-space control commands within distinct subspaces, as defined in Section~\ref{sec:subspace}. This framework utilizes two policies learned and executed in parallel: the $\mathcal{S}_K$-\textit{policy}, responsible for kinematic subspace force control, and the $\mathcal{S}_R$-\textit{policy}, responsible for redundant subspace motion control.

\textbf{Kinematic Force Control Policy}, $\mathcal{S}_K$-\textit{policy}, generates forces to control elements within $\mathcal{S}_K$, ensuring motion adheres to the object's kinematic constraints. For prismatic joints, feasible motion occurs along $\hat{Z}_o$, represented as $\{z\} \in \mathcal{S}_K$. The generated force $\mathbf{F}$ is calculated as:
\[
\mathbf{F}_{\text{prismatic}} = F \frac{\mathbf{v}}{\|\mathbf{v}\|},
\]
where $\mathbf{v}$ is the velocity vector parallel to $\hat{Z}_o$.

For revolute joints, feasible motion involves rotation around $\hat{Z}_o$, represented as $\{x, y\} \in \mathcal{S}_K$. The corresponding force is given by:
\[
\mathbf{F}_{\text{revolute}} = F \frac{\boldsymbol{\omega} \times \mathbf{r}}{\|\boldsymbol{\omega} \times \mathbf{r}\|},
\]
where $\boldsymbol{\omega}$ is the angular velocity vector, and $\mathbf{r}$ is the position vector relative to the axis $\hat{Z}_o$.

In both cases, the magnitude of $F$ must adaptively change based on the object's dynamics, which are often unknown and cannot be estimated beforehand. This adaptability is essential for achieving stable control during manipulation tasks. As previously discussed, adopting force control for the kinematic subspace is advantageous, as this subspace directly governs manipulation based on the object's kinematic configuration.

The primary objective of the $\mathcal{S}_K$-\textit{policy} is to regulate the velocity of the object’s joint to achieve a desired target velocity. While the direction of the applied force is dictated by the object’s kinematic constraints, the magnitude is learned and adaptively adjusted to accommodate variations in the object’s dynamics.

\textbf{Redundant Motion Control Policy}, $\mathcal{S}_R$-\textit{policy}, is responsible for generating motion within the redundant subspace, $\mathcal{S}_R$. This policy enables task optimizations such as avoiding obstacles, minimizing joint stress, and improving energy efficiency.

Figure~\ref{fig:door_simple} illustrates the advantages of leveraging the redundant subspace during task execution. By utilizing redundancy instead of imposing unnecessary pose constraints during trajectory planning, the manipulator’s dexterity can be fully exploited, resulting in more effective task execution. Despite its potential, many prior studies on articulated object manipulation have overlooked the redundant subspace, treating it as part of geometric constraints \cite{berenson2011task,jang2023motion,karayiannidis2016adaptive}, primarily due to the absence of a clear framework for planning trajectories in this subspace.

The $\mathcal{S}_R$-\textit{policy} aims to generate motion in $\mathcal{S}_R$ with the primary objective of maintaining the task for as long as possible. This simple yet effective goal indirectly alleviates joint stress and avoids collisions by strategically leveraging the redundant subspace. 

By separating policies, each subspace has a single objective, enabling focused control of specific goals.

The MDPs of each policy are as follows:

\subsubsection{State}
\label{sec:state}
The $\mathcal{S}_R$-\textit{policy} and $\mathcal{S}_K$-\textit{policy} utilize the same set of observations, which include the robot's joint angles $q \in \mathbb{R}^k$ and the actuated torque of the robot joints $\tau \in \mathbb{R}^k$, where $k$ denotes the robot's degrees of freedom. Additionally, the transformation matrix of the robot's end-effector, $\prescript{O}{}{T_E} \in SE(3)$, and the angular velocity or velocity of the object's joints $\omega, v$ are also shared between the policies. 

In the real world, the exact velocity of the object during manipulation cannot be measured directly. Instead, $\omega$ and $v$ were estimated by tracking the position of the robot's end-effector as $\hat{\omega}$ and $\hat{v}$. To satisfy the \textit{Markov Property} \cite{sutton2018reinforcement}, observations accumulated over 10 steps were used as the state. An LSTM \cite{hochreiter1997long} was employed as the feature extractor.
 
\subsubsection{Action}

The action of the $\mathcal{S}_K$-\textit{policy} is a discretized delta force, $\triangle F \in \mathbb{I}^4$. The delta force is selected from discrete values $\{0.1, 0, -0.1, 1\}$. This discretization helps stabilize the robot’s motion while maintaining a high control frequency (100 Hz), which is critical for adaptive force control. 

The action of the $\mathcal{S}_R$-\textit{policy} is the acceleration of motion in the redundant subspace, $\ddot{x}_R \in \mathbb{R}^n$, where $n$ represents the dimension of $\mathcal{S}_R$. 

\subsubsection{Reward}
The reward for \(\mathcal{S}_K\)-policy is:
\[
\mathcal{R}_{\mathcal{S}_K} = 
\begin{cases}
    1, & \text{if in the desired (angular) velocity} \\
    0, & \text{otherwise}
\end{cases}
\]
The reward for \(\mathcal{S}_R\)-policy is:
\[
\mathcal{R}_{\mathcal{S}_R} = 1 - k_1 ||\triangle \ddot{x}_R||_1 - k_2 \sum \log(-c_F).
\]
\(c_F\) denotes contact force due to collisions other than grasping. The second term reduces oscillation or unnecessary motion, while the last term encourages avoiding collisions. \(k_1\) and \(k_2\) are weights for the reward terms, set to 1 and 0.1 respectively. The desired angular velocity varies depending on the object, for instance, the desired angular velocities of valves are 0.7-0.8 rad/s, doors are 0.1-0.15 rad/s and drawers are 0.4-0.5 m/s.

\subsubsection{Terminal Conditions}
The episode terminates if the robot exceeds the joint limit boundary, loses contact with the object, or the designated manipulation time is over. Losing the object is defined as the loss of contact detected in the robot's gripper. Terminal rewards of \(-100\) are applied for conditions other than the time restriction. 

\subsection{Dynamics for Hybrid Force and Motion Control}
The action of the $\mathcal{S}_K$-\textit{policy}, $\triangle F$, adjusts the desired magnitude of the force vector in $\mathcal{S}_K$ as $F_{des} \leftarrow F + \triangle F$. The resulting $F_{des}$ is then multiplied by the unit force vector in $\mathcal{S}_F$ to compute the desired force vector $\mathbf{F}_{des}$.

Simultaneously, the action of the $\mathcal{S}_R$-\textit{policy}, $\ddot{x}_R$, is integrated with the motion planner output from the geometric subspace ($\dot{x}_G$, $x_G$) to generate the desired velocity $\dot{x}_{des}$ and position $x_{des}$. This integration allows coordinated motion control across the geometric and redundant subspaces.

To ensure independent force and motion control, a selection matrix $S$ is defined by a task space decomposition. The matrix $S$ specifies the elements in the task space assigned to motion control, corresponding to the geometric subspace $\mathcal{S}_G$ and the redundant subspace $\mathcal{S}_R$. This masks all other elements for motion control. Conversely, the complementary matrix $(I - S)$ is applied to the force control term, ensuring that force control is exclusively applied to the kinematic subspace $\mathcal{S}_K$. This partitioning guarantees that motion and force control operate independently within their designated subspaces.

\subsection{Enhancing Sample Efficiency}
Although we decomposed the action space into respective subspaces for articulated object manipulation through task space decomposition, contact-rich manipulation using a manipulator robot requires considerable learning time due to numerous constraints. We utilized $\mathcal{S}_R$ as a geometric constraint and employed a manual method for adjusting the force gain based on specific conditions (refer to Section \ref{sec:manual}) to create a dataset for training \cite{ball2023efficient}. The training utilized a 1:1 ratio of online data collected through exploration and offline data pre-collected. This method was specifically applied to the challenging valve manipulation task, which involves diverse positions, orientations, and radii.

\section{EXPERIMENT} \label{sec:experiment}
Through experiments, we verified that SwRL enhances manipulation performance across various articulated objects. The objects used in our experiments include a handwheel valve, a lever handle valve, a door, and a drawer. The common goal of each task is to open or turn the object as far as possible while adhering to joint limits, avoiding singularity poses, and preventing self-collision and collision with other objects. Valves are the primary focus of our experiments.
To simulate practical real-world settings, we varied not only the position, orientation, and size of the objects but also the joint friction parameters to create a more realistic environment. 
\subsection{Technical Details}
We built a simulation environment to train and test our method using MuJoCo \cite{todorov2012mujoco}. The high-level controller is designed in C++ and integrated with the Python environment for the training and testing phases using pybind11 \cite{pybind11}. We utilize a Franka Research 3 robotic arm equipped with a parallel gripper, operating at a control frequency of 1 kHz. For the RL algorithm, we implemented the truncated quantile critic (TQC) \cite{kuznetsov2020controlling}. RL policies operate at a frequency of 100 Hz and are interpolated during the hybrid force/motion controller. The training was performed on a computer with the following specifications: 13th generation Intel Core i7-13700F CPU with 32GiB of RAM and an NVIDIA RTX 4070 Ti GPU.

\subsection{Baselines} 
We compared SwRL with four baselines.
\begin{itemize}
    \item $\textit{Manual}$: \label{sec:manual}As observed in many previous studies \cite{berenson2011task,jang2023motion,karayiannidis2016adaptive}, the manual method treats redundant subspace elements as geometric constraints. In our experiments, the manual method maintains the redundant subspace elements from the initial grasping posture, creating motions that consider the redundant space as a geometric constraint. For the force magnitude, $F$ discretely changes based on the desired angular velocity range of the object and the joint limit boundary of the robot.
    \item \textit{Behavioral Cloning}: The agent receives inputs as described in Section \ref{sec:manual}. We employ an LSTM feature extractor, which is the same condition as our method, followed by a two-layer MLP. The agent imitates the dataset collected using the manual approach detailed in Section \ref{sec:manual}.
    \item \textit{Reinforcement Learning}: We also compare SwRL with the independent actor-critic method to a vanilla RL approach, specifically a single actor-critic method. With all other parameters kept constant, only the action spaces are concatenated. The agent is trained with an identical reward function, summing $\mathcal{R}_{\mathcal{S}_R}$ and $\mathcal{R}_{\mathcal{S}_K}$. This setting obscures the task space decomposition, highlighting the discrepancy between using task space decomposition and not using it.
\end{itemize}
To further evaluate how each policy impacts overall performance, we compared the results of using only the $\mathcal{S}_R$-policy and the $\mathcal{S}_K$-policy independently.

\subsection{Metrics} 
We evaluate our method against baselines using three metrics:
\begin{itemize}
    \item $\textit{Articulation Position}:$ This metric measures the average joint articulation of the object over 120 cases, varying the object's physical parameters such as friction loss and pose.
    \item $\textit{Relative Articulated Percentage}:$ RMP quantifies the improvement in manipulation performance compared to the manual method. It is calculated as follows:\\
    \[\text{RMP} = \frac{\theta_{\text{method}} - \theta_{\text{manual}}}{\theta_{\text{manual}}}\times100.\] 
    \item $\textit{Manipulability}:$ Manipulability measures how well the robot can control its end-effector's position and orientation in different directions. We adapt Yoshikawa's manipulability index $\omega$ \cite{yoshikawa1985manipulability} and visualize the variation of manipulability according to the joint articulation of the object based on the method $\theta_{method}$. The manipulability index is defined as $\omega=\sqrt{\text{det}(JJ^T)}$.
\end{itemize}
\begin{figure}[htp!]
\centering
        \subfloat[\centering {Handwheel valve task}]{
        \includegraphics[width=.65\textwidth]{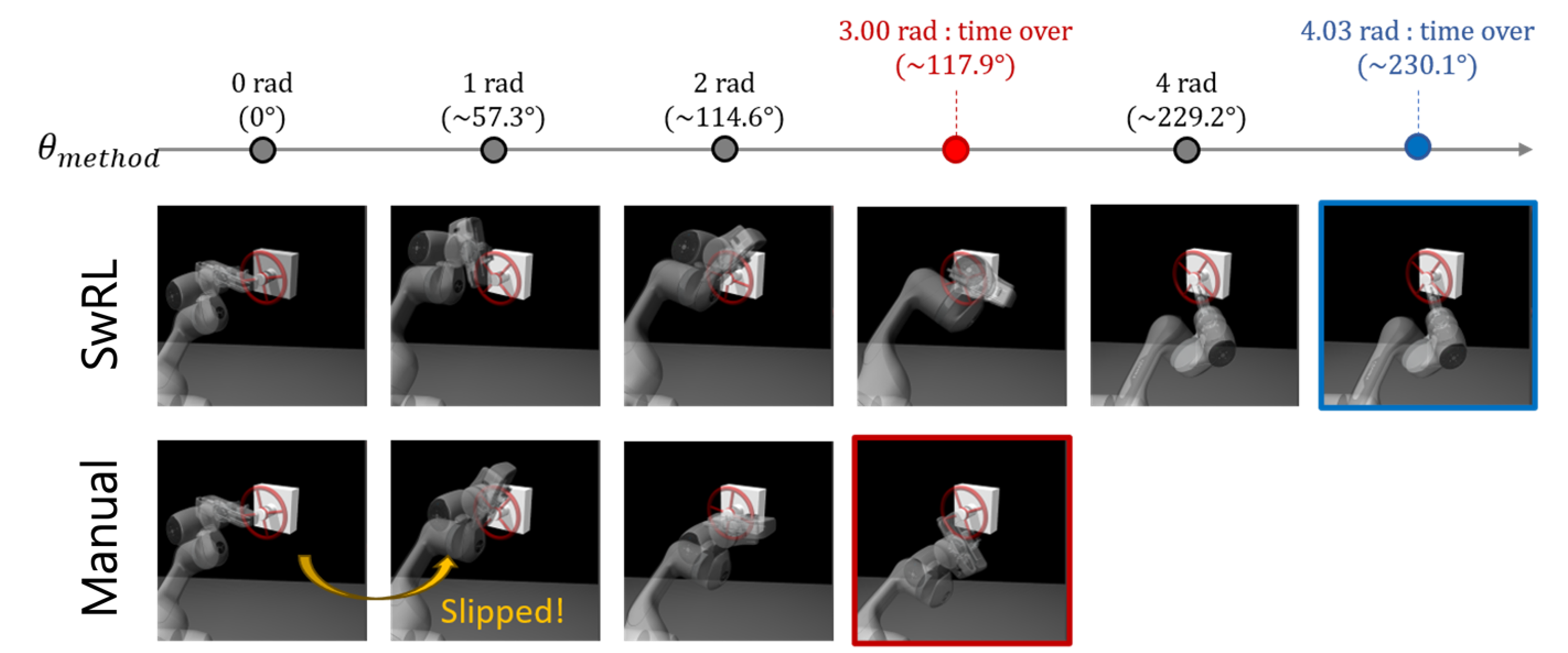}
        \label{fig:ss_handle}}
        \\ \hspace{10pt}
        \subfloat[\centering {Lever handle valve task}]{
        \includegraphics[width=.65\textwidth]{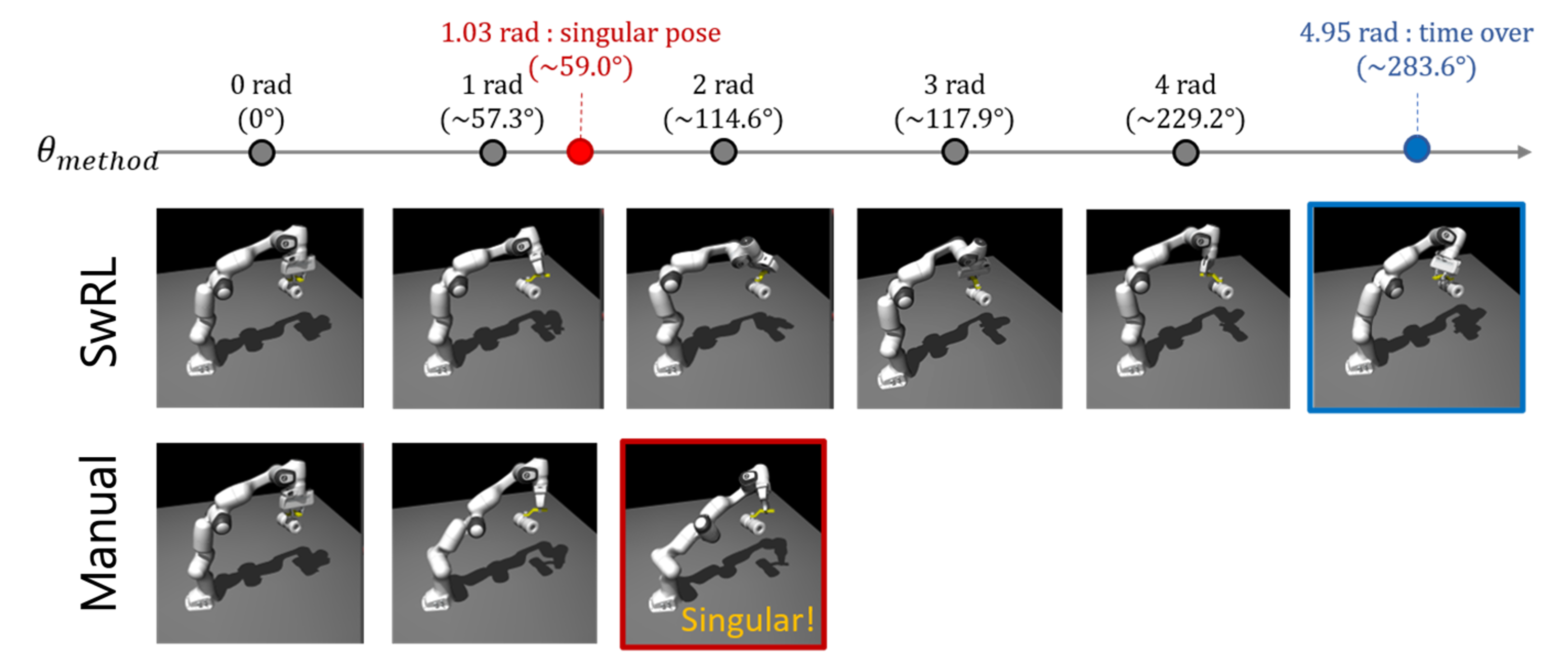}
        \label{fig:ss_valve}
        }
        \\ \hspace{10pt}
        \subfloat[\centering {Door task}]{
        \includegraphics[width=.65\textwidth]{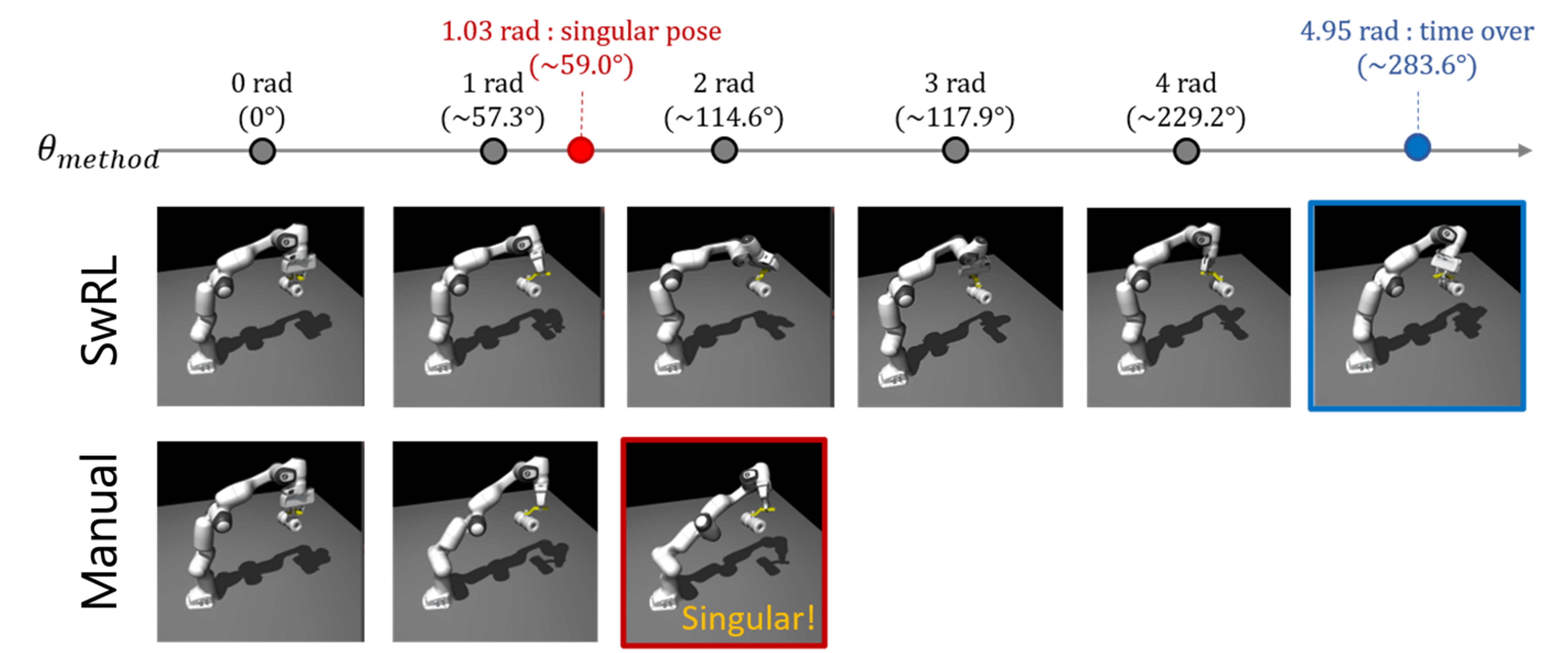}
        \label{fig:ss_door}
        }
        \\ \hspace{10pt}
        \subfloat[\centering {Cabinet task}]{
        \includegraphics[width=.65\textwidth]{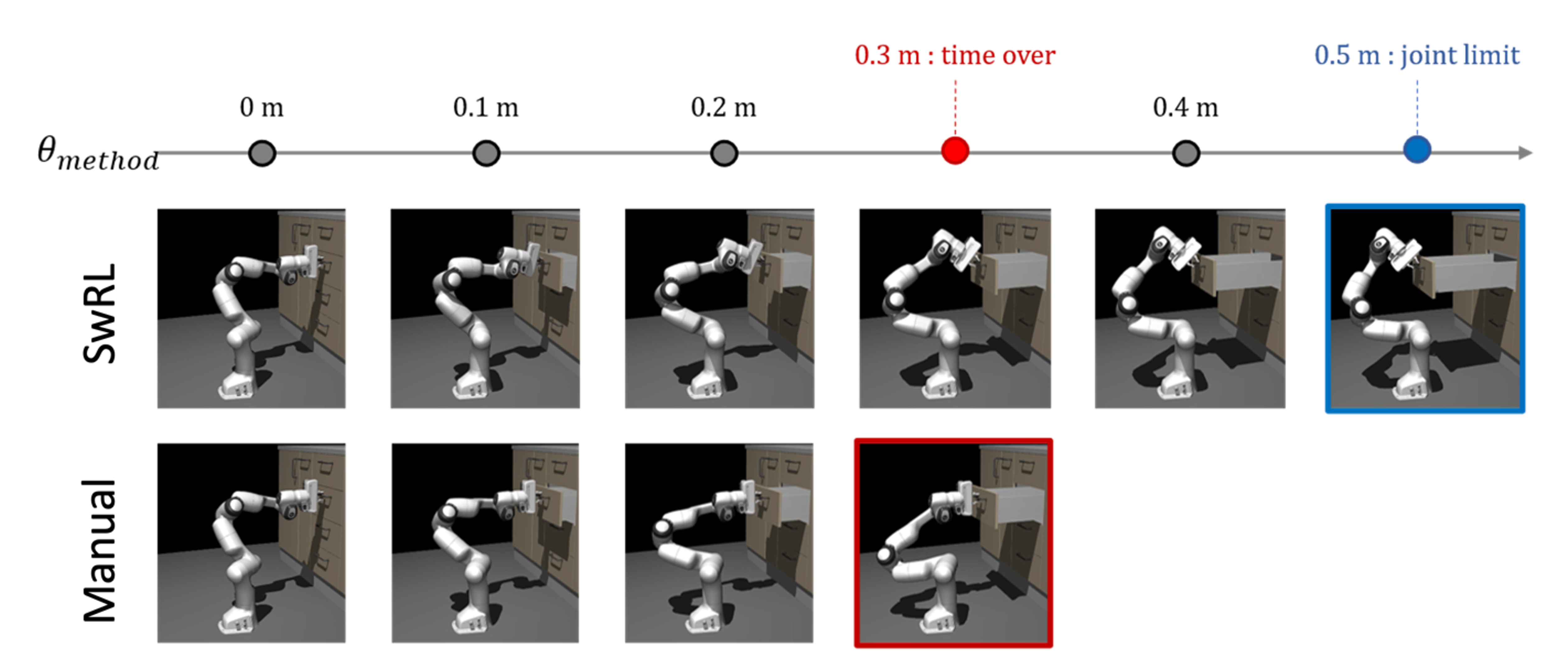}
        \label{fig:ss_cabinet}
        }
    \caption{Comparison of SwRL and manual method across four tasks. The blue boxes and labels indicate the terminal pose of the end-effector and the resulting joint angle of the object using the proposed method. The red boxes and labels indicate the same for the manual approach under identical environment settings.}
    \label{fig:ss}
\end{figure}
\newpage

\subsection{Simulation Result}

\(\textit{During Training:}\) 
We compare the episodic return during the training process of SwRL and the vanilla RL method, which does not utilize task space decomposition. While the vanilla RL method trains a single policy, we logged the reward functions $\mathcal{R}_{\mathcal{S}_K}$ and $\mathcal{R}_{\mathcal{S}_R}$ independently to compare the accumulated rewards with SwRL.

\begin{figure}[t!]
    \centering
        \subfloat[\centering {Handwheel valve task}]{
        \includegraphics[width=.33\textwidth]{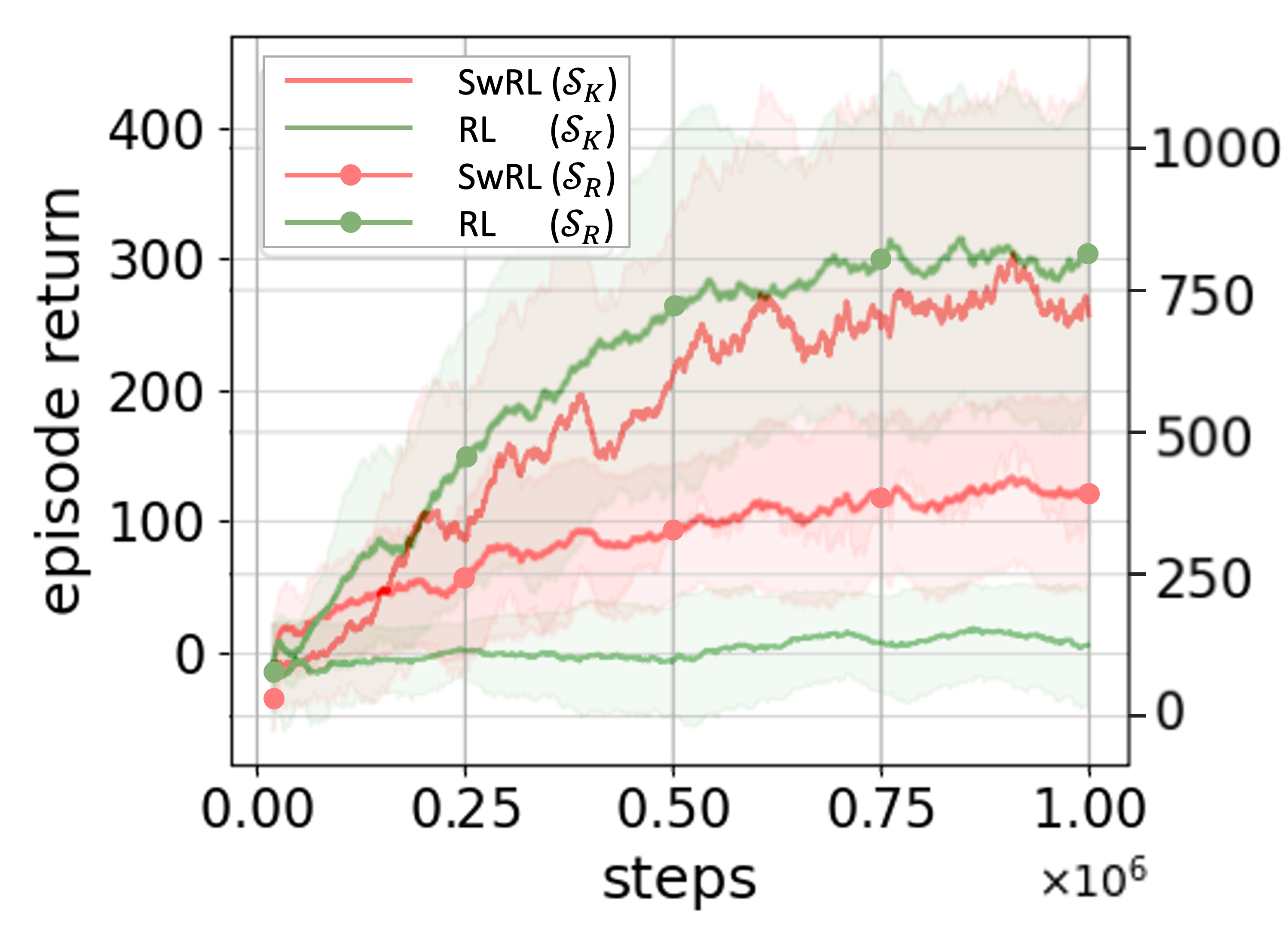}
        \label{fig:handle_reward}}
        \hspace{10pt}
        \subfloat[\centering {Lever handle valve task}]{
        \includegraphics[width=.33\textwidth]{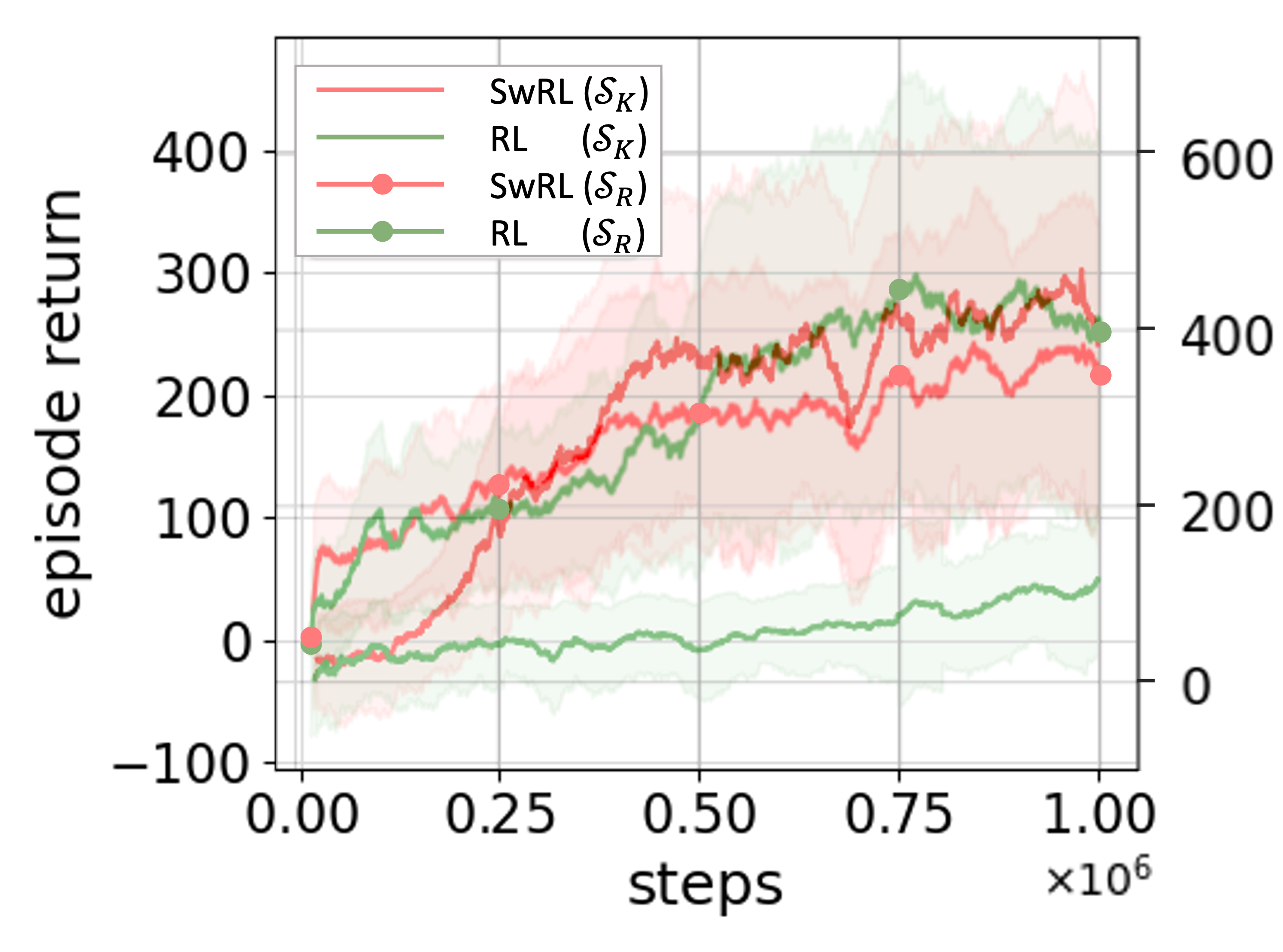}
        \label{fig:valve_reward}
        }\hspace{10pt}
        \subfloat[\centering {Door task}]{
        \includegraphics[width=.33\textwidth]{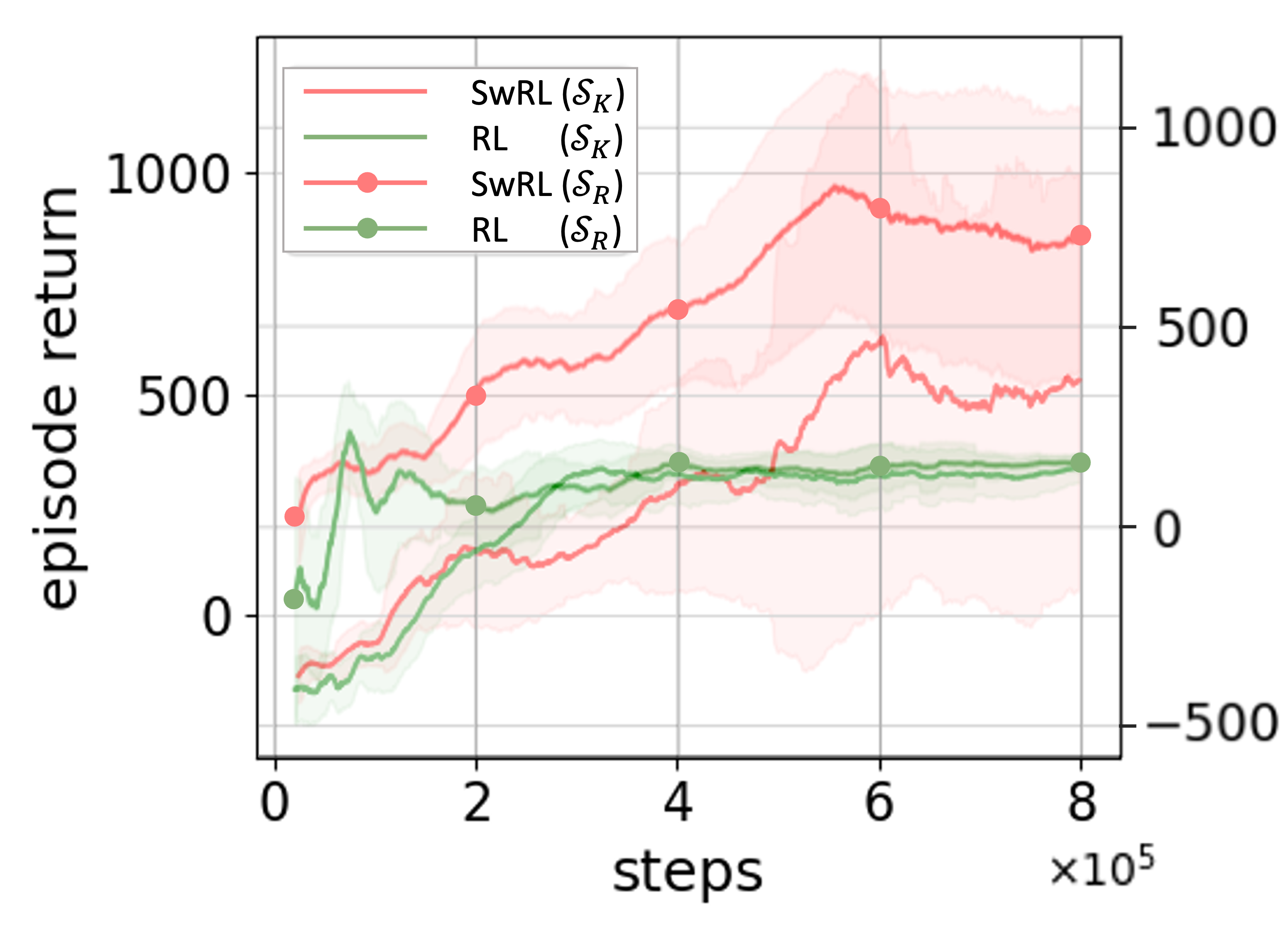}
        \label{fig:door_reward}
        }
        \hspace{10pt}
        \subfloat[\centering {Drawer task}]{
        \includegraphics[width=.33\textwidth]{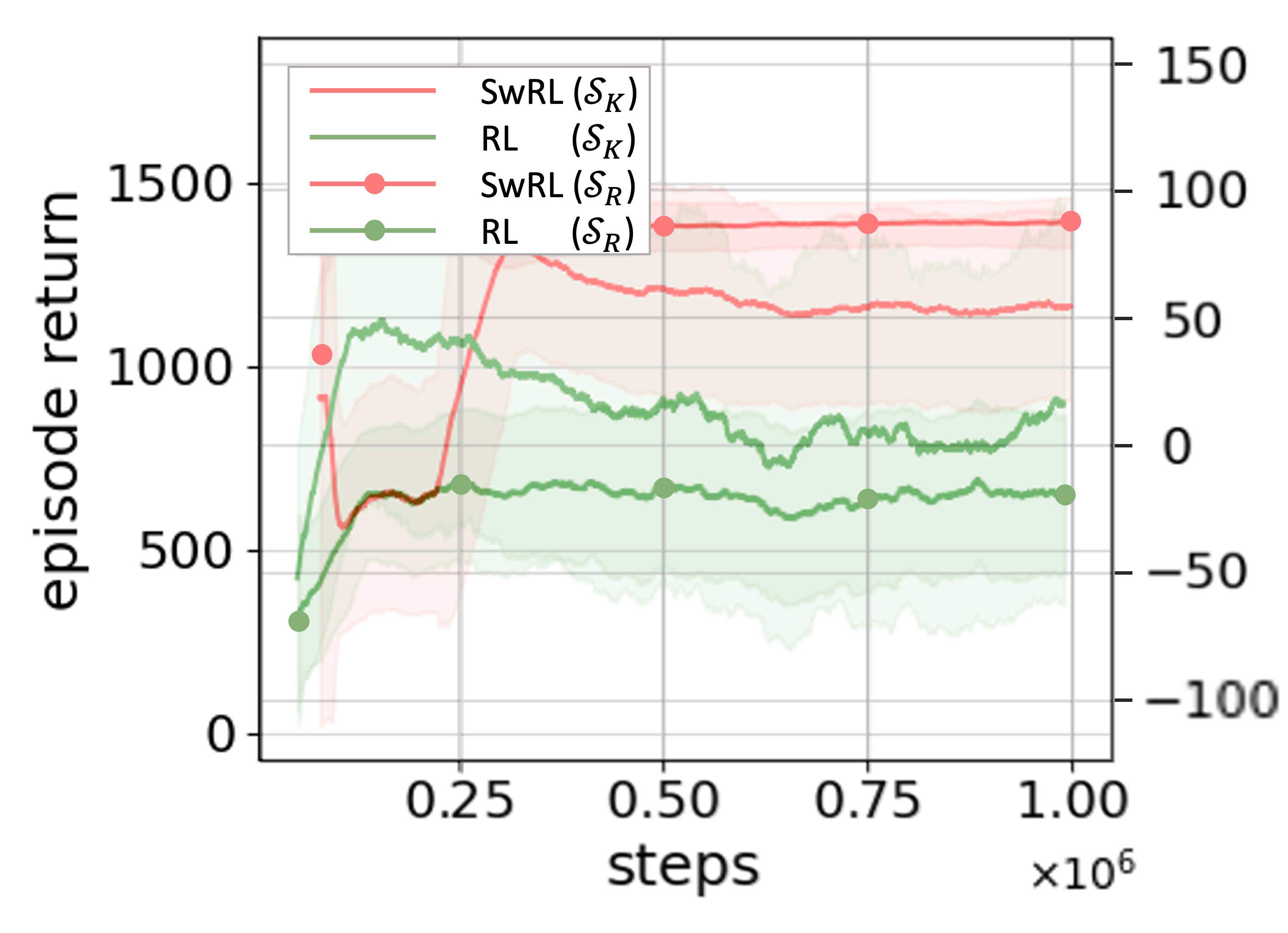}
        \label{fig:drawer_reward}
        }
        
    \caption{\label{fig:reward_graph}{The episode return during the training process for each task is illustrated. The \textbf{left y-axis} corresponds to \textbf{SwRL} (\(\mathcal{S}_K\)) and \textbf{RL} (\(\mathcal{S}_K\)), while the \textbf{right y-axis} corresponds to \textbf{SwRL} (\(\mathcal{S}_R\)) and \textbf{RL} (\(\mathcal{S}_R\)). Note that, the episode return for $\mathcal{S}_K$ reflects articulation of the object, whereas the episode return for $\mathcal{S}_R$ corresponds to the episode length, as it increases with the episode duration. The optimal result is when both graphs show a high accumulated return.}}
\end{figure}

The return from the $\mathcal{S}_K$-\textit{policy} reflects successful object articulation, whereas the return from the $\mathcal{S}_R$-\textit{policy} correlates with episode duration. For effective articulated object manipulation, balanced learning of both force control and redundant space motion is crucial to achieving high returns. Figure~\ref{fig:reward_graph} demonstrates that the proposed method achieves faster convergence and higher returns for the $\mathcal{S}_K$-\textit{policy} across all tasks.

An exception is observed in the handwheel valve task (Figure~\ref{fig:handle_reward}), where SwRL shows a lower return for the $\mathcal{S}_R$-\textit{policy} compared to the vanilla RL method. However, the $\mathcal{S}_K$-\textit{policy} return for the vanilla RL method remains significantly lower. This indicates that the vanilla RL method struggles with force control, often resulting in the robot remaining stationary and failing to articulate the object, which in turn inflates the reward for the $\mathcal{S}_R$-\textit{policy} due to prolonged episode duration.

These results highlight that the proposed task space decomposition enables balanced learning between force control and redundant motion, significantly improving training efficiency and overall performance.

\(\textit{After Training:}\)

\begin{table*}[b!]
\centering
\caption{Average Articulated Joint Position and Relative Articulated Percentage\label{tab:average_turn_rmp}}
\begin{adjustbox}{max width=\textwidth}
\begin{tabular}{c|c c c c c c|c c c c c c}
\hline
\multirow{2}{*}{Object} & \multicolumn{6}{c|}{\textit{Average Articulated Joint Position}} & \multicolumn{6}{c}{\textit{Relative Articulated Percentage}} \\ \cline{2-13} 
                        & SwRL & Manual & SwRL-$\mathcal{S}_R$ & SwRL-$\mathcal{S}_K$ & RL & BC & SwRL & Manual & SwRL-$\mathcal{S}_R$ & SwRL-$\mathcal{S}_K$ & RL & BC \\ \hline
Handwheel valve         & \(\mathbf{258.6^\circ}\) & \(202.6^\circ\) & \(217.7^\circ\) & \(232.2^\circ\) & \(31.78^\circ\) & \(214.6^\circ\) & \(\mathbf{31.5\%}\) & \(0.0\%\) & \(15.4\%\) & \(21.3\%\) & \(-79.3\%\) & \(13.3\%\) \\ \hline
Lever handle valve      & \(\mathbf{173.3^\circ}\) & \(159.8^\circ\) & \(170.4^\circ\) & \(155.8^\circ\) & \(99.0^\circ\) & \(129.3^\circ\) & \(\mathbf{17.6\%}\) & \(0.0\%\) & \(11.8\%\) & \(2.7\%\) & \(-30.0\%\) & \(-2.8\%\) \\ \hline
Door                    & \(41.3^\circ\)  & \(36.7^\circ\)  & \(\mathbf{44.7^\circ}\)  & \(32.1^\circ\)  & \(9.16^\circ\) & \(33.2^\circ\)  & \(26.3\%\) & \(0.0\%\) & \(\mathbf{36.8\%}\) & \(-1.7\%\) & \(-71.9\%\) & \(1.7\%\) \\ \hline
Drawer                  & \(\mathbf{0.503m}\)      & \(0.30m\)       & \(0.098m\)      & \(0.314m\)      & \(0.227m\)     & \(0.169m\)      & \(\mathbf{67.5\%}\) & \(0.0\%\) & \(-67.3\%\) & \(4.67\%\) & \(-24.3\%\) & \(-43.67\%\) \\ \hline
\end{tabular}
\end{adjustbox}
\end{table*}

Table \ref{tab:average_turn_rmp} shows quantitative results over 4 objects. For the handwheel valve and lever handle valve, 120 cases are executed with varying positions, orientations, sizes, and friction loss. For the door, friction loss varies for 10 cases, and for the drawer, 10 cases are randomly executed, opening the 1st to 4th floor of the drawer. The result shows that SwRL outperforms all baselines over every object in manipulation. 
In Table \ref{tab:average_turn_rmp}, $\textit{Average Articulated Joint Position}$ presents the average angle of manipulation for each object, and $\textit{Relative Articulated Percentage}$ shows the relative performance improvement over the manual method, considering the varying absolute manipulability due to the relative positions of the object and the robot. To prevent overestimation due to outliers, these values were clipped between \(-100\) and \(+100\). A value of \(+100\) indicates that the manipulation angle achieved is more than twice that of the manual method.

\begin{figure}[t!]
    \centering
        \subfloat[\centering {Manipulability measure for handwheel valve.}]{
        \includegraphics[width=.3\textwidth]{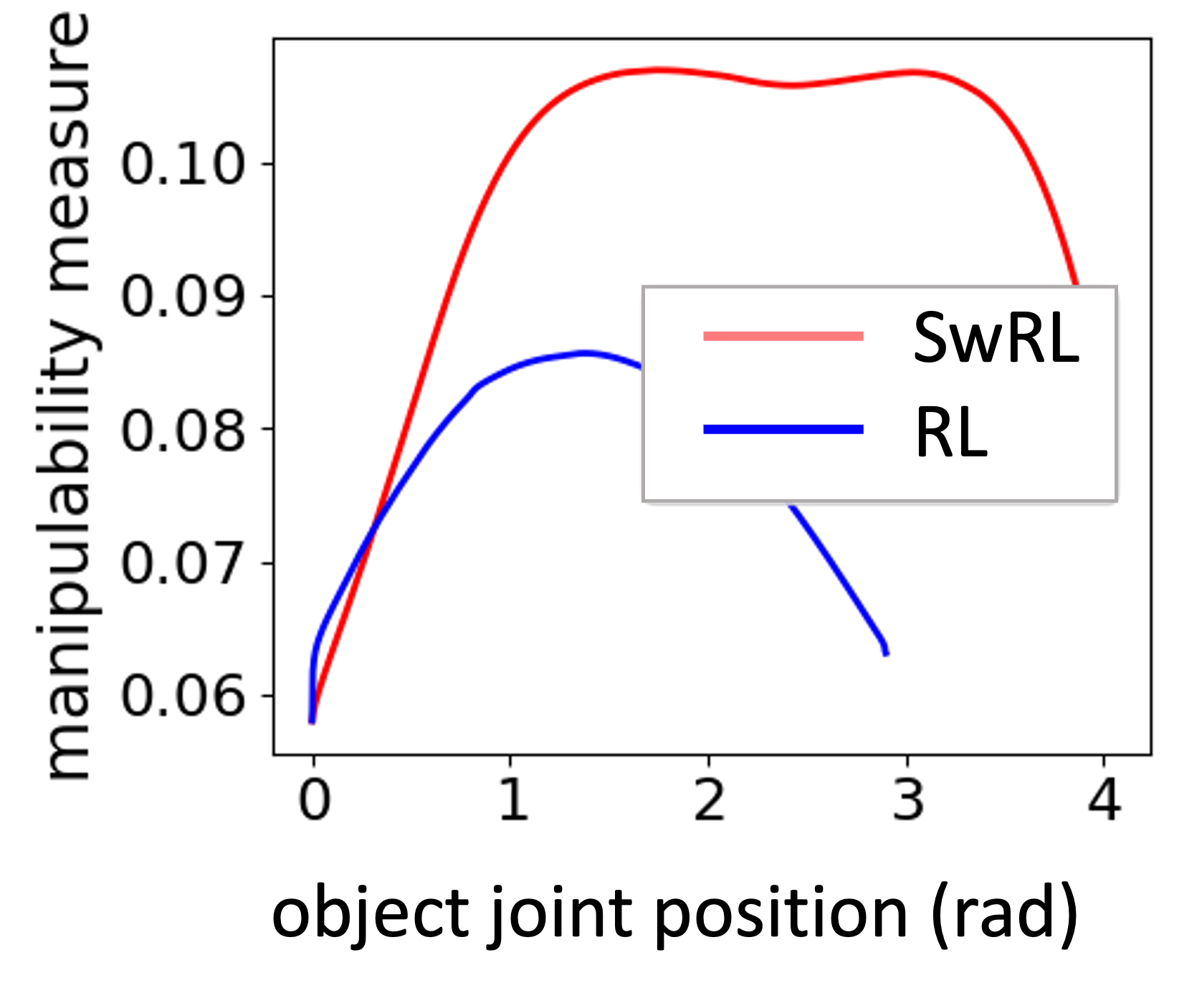}
        \label{fig:manipulability_handle}}
        \subfloat[\centering {Manipulability measure for lever handle valve.}]{
        \includegraphics[width=.3\textwidth]{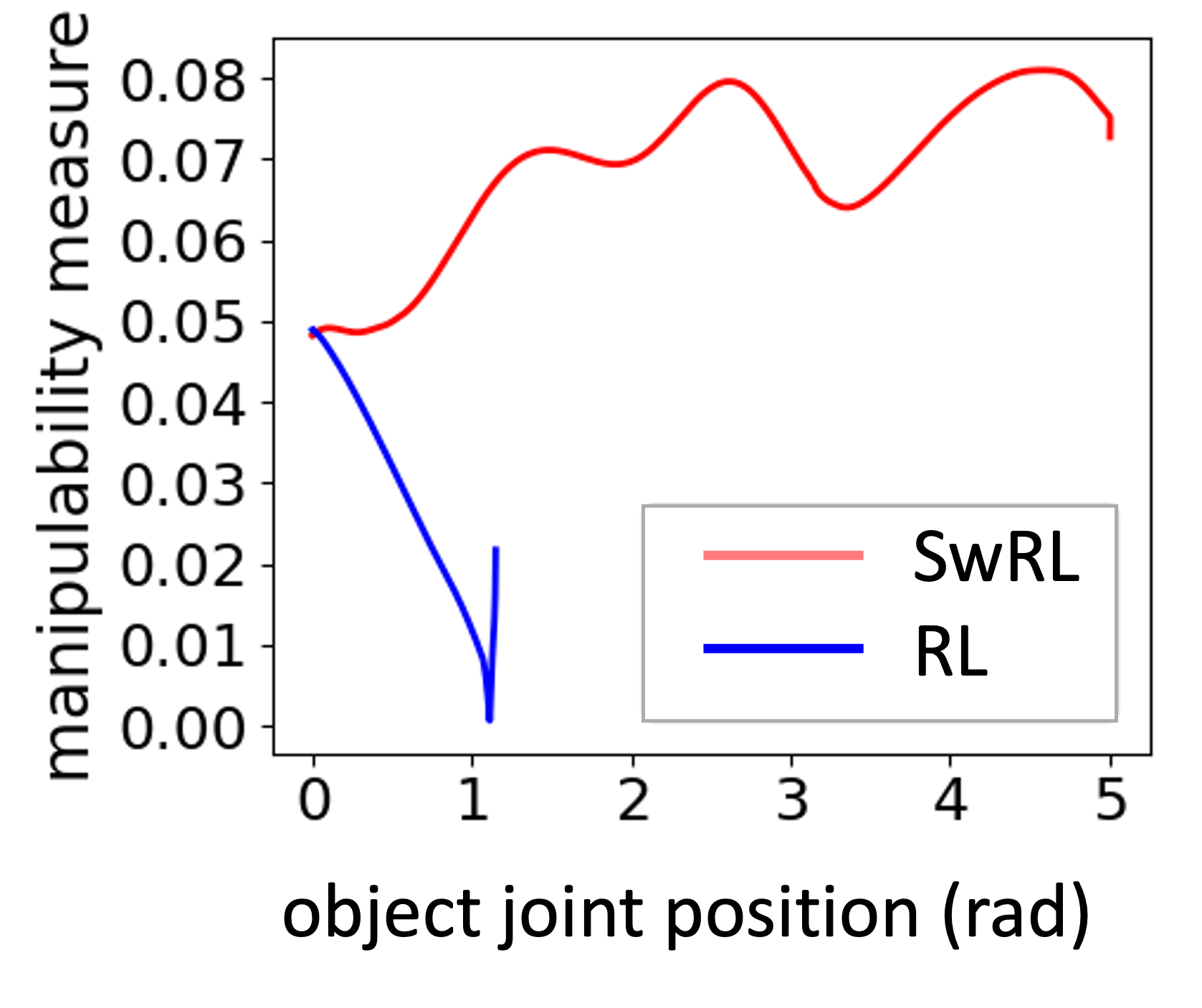}
        \label{fig:manipulability_valve}
        }
        \hspace{10pt}
        \subfloat[\centering {Manipulability measure for door case.}]{
        \includegraphics[width=.3\textwidth]{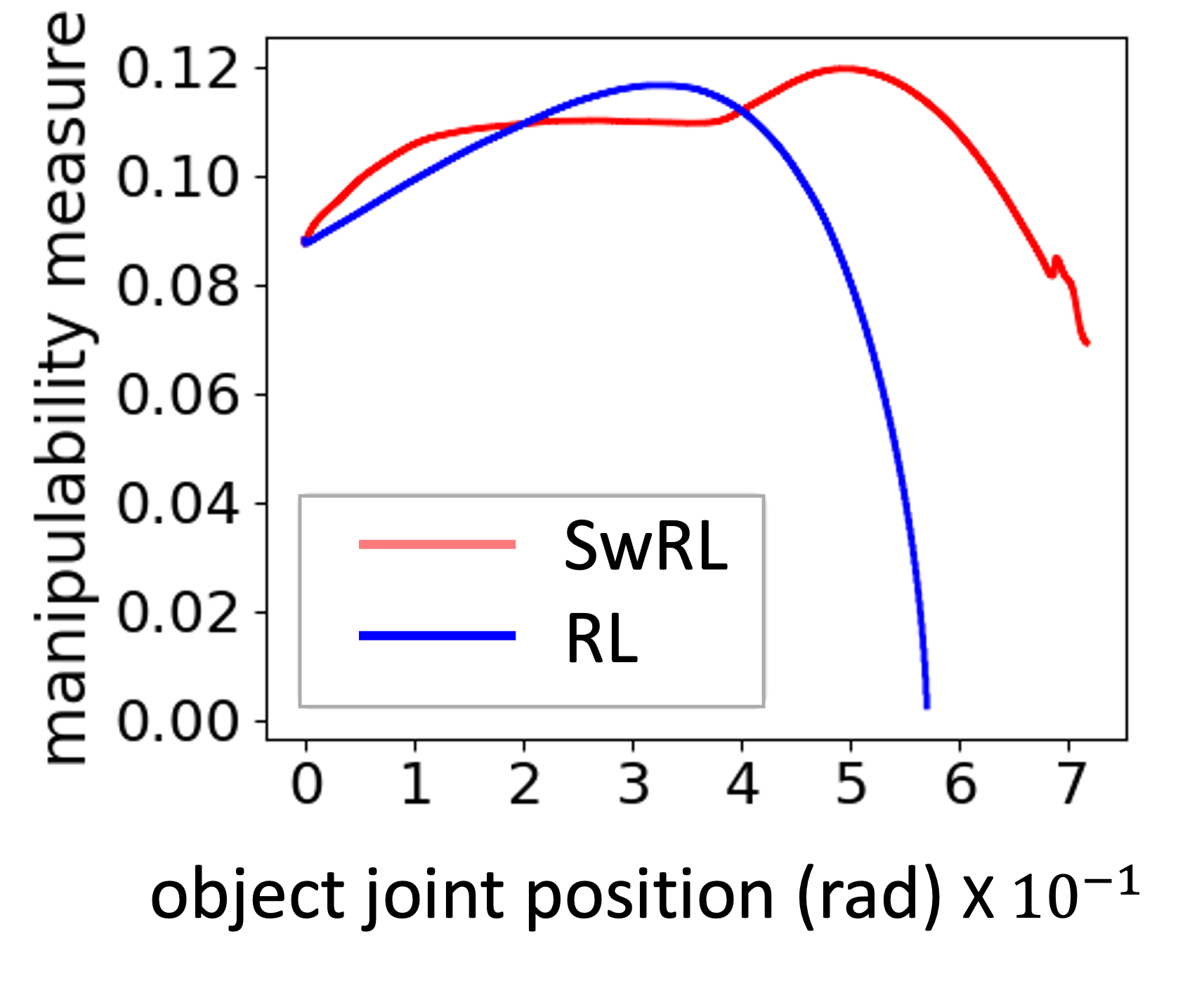}
        \label{fig:manipulability_door}
        }
        \subfloat[\centering {Manipulability measure for cabinet case.}]{
        \includegraphics[width=.3\textwidth]{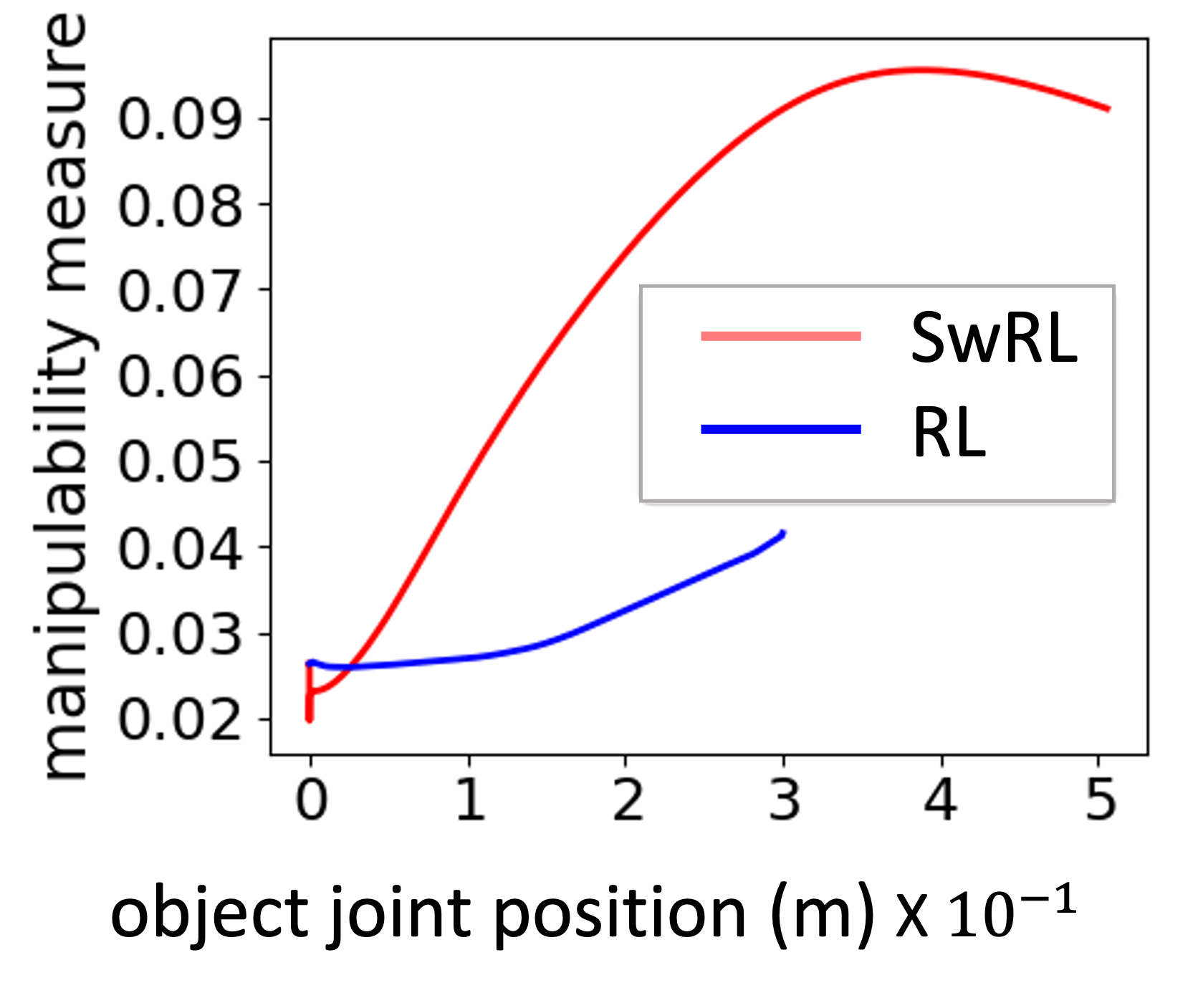}
        \label{fig:manipulability_cabinet}
        }
    \caption{\label{fig:manipulability}{Comparison of the manipulability measure between the proposed method and the manual approach as a function of the change of the object's joint position.}}
\end{figure}

As observed in Tables \ref{tab:average_turn_rmp}, SwRL outperformed the baseline methods in all cases. Figure \ref{fig:ss} captures one motion case per object from various random cases and Figure \ref{fig:manipulability} shows the changes in manipulability for each method at the corresponding object angles. In Figure \ref{fig:ss}(a), the manual method experienced grip slippage on the valve between 0 and 1 radian due to poor initial force control. The episode ended at 3 radians ($117.9^\circ$) near the joint limit. In contrast, SwRL gradually increased the force gain for stable gripping and utilized the redundant subspace to avoid the joint limit, achieving a rotation of 4.03 radians ($230.1^\circ$). The manipulability plot in Figure \ref{fig:manipulability_handle} indicates that our method maintained high manipulability throughout.

In Figure \ref{fig:ss}(b), the manual method encountered a singularity pose while satisfying geometric constraints. Conversely, SwRL used the redundant space to avoid singularity poses, achieving a rotation of 4.95 radians ($283.6^\circ$). Similarly, in Figure \ref{fig:ss}(c), the manual method resulted in a collision with the door due to a singularity pose during the opening motion. However, SwRL avoided collisions and maintained high manipulability as shown in Figure \ref{fig:manipulability_door}, allowing the door to be opened to 0.72 radians ($41.1^\circ$). In Figure \ref{fig:ss}(d), although our method failed to stop pulling the drawer before reaching the joint limit, it maintained a configuration with high manipulability as shown in Figure \ref{fig:manipulability_cabinet}, resulting in a longer length of the drawer opening. In all the cited examples, the robot's speed converged to zero when further progress was not possible, preventing collisions or reaching joint limits.

\subsection{Real World Implementation}
\begin{figure}[h!]
    \centering
    \includegraphics[width=0.5\textwidth]{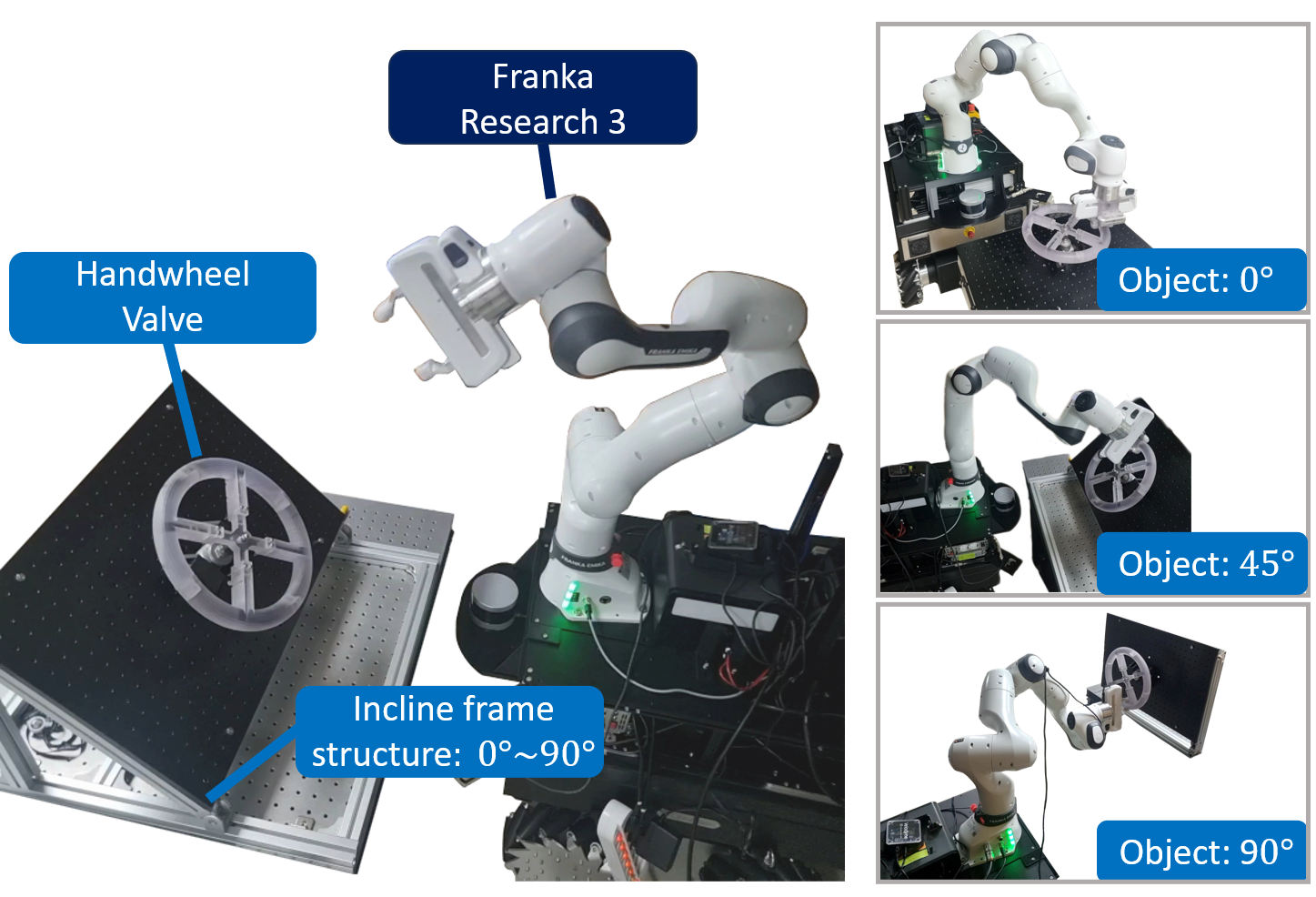}
    \caption{Real-world experiment settings on three different valve configurations.}
    \label{fig:realworld}
\end{figure}
\begin{figure}[h!]
    \centering
    \includegraphics[width=0.6\textwidth]{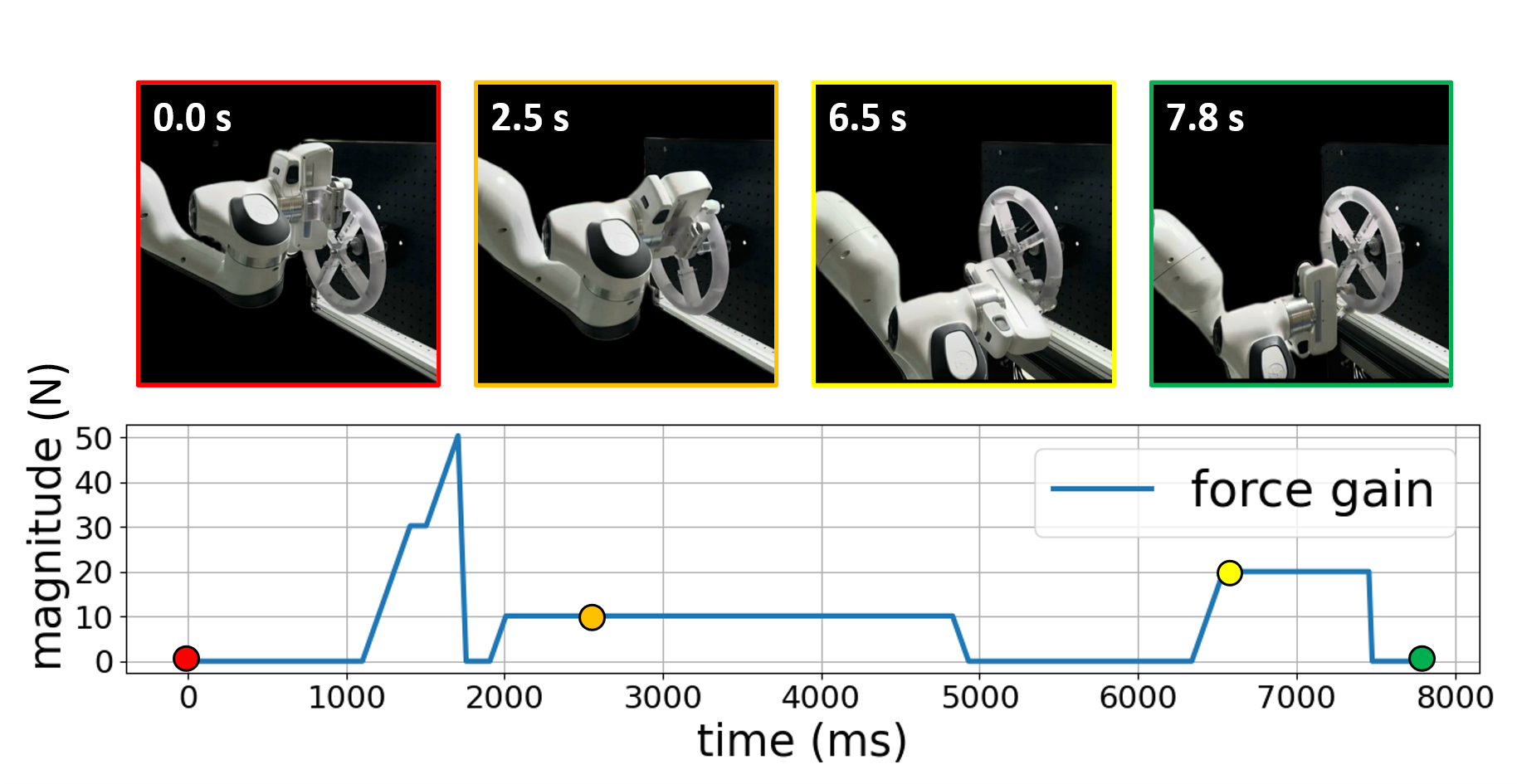}
    \caption{Valve turning motion and corresponding force profile with an unseen real-world valve.}
    \label{fig:force_profile}
\end{figure}

Our method is validated in a real-world environment. The policy trained in simulation is implemented on a real robot. The task involves turning a handwheel valve in three different poses as shown in Figure \ref{fig:realworld}. Among the observation spaces, the angular velocity of the object is estimated through the positional displacement of the end-effector.
Figure \ref{fig:force_profile} provides a snapshot of the end-effector motion and the corresponding force gain for the valve at $90^\circ$. From 1s to 2s, the force rapidly peaks to 50N to overcome the static friction of the valve, then decreases and maintains a constant force to keep the valve rotating. Although there was a stopping period from 5s to 6s, the motion recovered to a rotating state up to the maximum range of the joint limit. 

Given that physical parameters such as the valve's friction are unknown in the real-world setting, the suggested method effectively modulated the force to rotate the valve as in the simulation. The \(\mathcal{S}_K\)-policy was trained to maintain the desired angular velocity of the object regardless of various physical parameters. The \(\mathcal{S}_R\)-policy also demonstrated the utilization of redundant space motion to avoid joint stress. The results show that policies trained using task space decomposition effectively cope with unseen real-world settings. Additional motions in various cases are shown in our video due to volume limitations.

\subsection{Comparison with Planning Method}
To compare our method with a planning-based approach, we implemented one of the sampling-based planning algorithms, CBiRRT (Constrained Bi-directional Rapidly-exploring Random Tree) \cite{berenson2009manipulation}. CBiRRT introduces the concept of a Task Space Region (TSR), which represents the geometrically or kinematically constrained space of the end-effector. It generates constrained motion by projecting randomly sampled joint configurations onto this region.

We applied CBiRRT to valve-turning and door-opening tasks and compared its performance with our proposed method. Note that CBiRRT plans motion trajectories for all task space dimensions, utilizing constrained space in $\mathcal{S}_K$ and $\mathcal{S}_G$.

In the valve-turning task, SwRL achieved $272.2^\circ$ with an RMP of 168.1\%. Using the CBiRRT method, the results were $297.9^\circ$ with an RMP of 193.8\% and planning time of 380.56 seconds, averaged over 10 planning attempts. For the door-opening task, our method achieved $41.3^\circ$ with an RMP of 26.3\%, whereas the CBiRRT method achieved $32.0^\circ$ with an RPM of $-$12.8\% planning time of 0.42 seconds, averaged over 10 planning attempts. The CBiRRT method showed better articulated joint motion for the valve-turning task compared to the proposed method but performed worse in the door-opening task.
This demonstrates that SwRL provides object manipulation performance comparable to that of offline planning. Moreover, since offline planning requires a long computation time and cannot respond in real-time to dynamic situations, the proposed method offers a significant advantage.

Furthermore, the planned path involved unnecessary and irregular motion that yielded repeated openings and closings of the door and valve. In the door-opening task, the CBiRRT method took between 15 to 30 seconds accordingly, while the proposed method completed the task in 8 seconds. Detailed robot motions can be seen in the supplementary video.

\section{CONCLUSIONS} \label{sec:conclusion}
In this study, we proposed SwRL, a framework for articulated object manipulation through task decomposition in an object-oriented frame. By decomposing the task space, we divided it into constrained and redundant subspaces and developed subspace-wise RL method with hybrid motion and force control.

Our experiments demonstrated that utilizing the redundant space in articulated object manipulation, where the robot's motions are constrained, enhances the robot's dexterity and improves manipulation performance. Additionally, by learning the force gain, the robot could adaptively respond to the unknown physical parameters of the objects. This approach outperforms methods that learn or plan the full task space of the robot end-effector for object manipulation, highlighting the advantages of subspace-wise learning.

The potential of our method extends beyond articulated object manipulation, as our framework can be applied to various types of manipulation tasks. Unconstrained motion can be represented as the full space of the redundant subspace, and goal-reaching can be decomposed into constrained and redundant subspaces. 

Our current approach requires that the decomposition of these subspaces be pre-determined. While this method can be applied consistently based on the type of the object's joint, it necessitates prior knowledge of the object and situation. To enhance the generalizability of our approach, future work will focus on developing a task space decomposition module that can adapt to various objects and scenarios, thereby integrating seamlessly with our framework.

\bibliographystyle{unsrt}  
\bibliography{references}
\end{document}